\pdfoutput=1
\documentclass[final,nopreprintline]{elsarticle}

\usepackage{subcaption}
\usepackage{xcolor}

\usepackage{amsmath}
\usepackage{amsfonts}
\usepackage{amsmath}
\usepackage{amsfonts}
\usepackage{amssymb}
\usepackage{bm}
\usepackage{physics}
\usepackage{nicefrac}
\usepackage{paralist}
\usepackage{enumitem}
\usepackage{multirow}
\usepackage{relsize}

\usepackage{tikz}
\usepackage{pgfplots}

\usepackage[protrusion=true,tracking=true,kerning=true,expansion,final]{microtype}      %
\usepackage{booktabs} %

\usepackage{varioref} %
\usepackage[breaklinks=true,colorlinks,bookmarks=true]{hyperref}
\usepackage[capitalize]{cleveref}
\crefname{appsec}{appendix}{appendices}
\Crefname{appsec}{Appendix}{Appendices}
\usepackage{url}

\biboptions{authoryear}

\newcommand{\eb}[1]{{\scriptsize\,$\pm$\,#1}}

\def\1{\bm{1}}

\def\eps{{\epsilon}}

\def\vh{{\bm{h}}}

\def\vx{{\bm{x}}}

\def\mW{{\bm{W}}}
\def\mX{{\bm{X}}}

\def\sR{{\mathbb{R}}}

\vbadness=\maxdimen
\renewcommand{\cite}[1]{\citep{#1}}
\definecolor{mydarkblue}{rgb}{0,0.08,0.45}
\definecolor{urlcolor}{rgb}{0,.145,.698}
\definecolor{linkcolor}{rgb}{.71,0.21,0.01}
\hypersetup{ %
	pdfkeywords={},
	pdfborder=0 0 0,     
    pdfpagemode=UseOutlines,
	colorlinks=true,
	bookmarksnumbered=true,
	urlcolor=urlcolor,
	linkcolor=linkcolor,
	citecolor=mydarkblue,
	filecolor=mydarkblue,
	pdfview=FitH}

\definecolor{darkgreen}{rgb}{0.0, 0.5, 0.0}

\newcommand{\noise}{\bm{\eta}}
\newcommand{\goodtype}{\emph{indicative}}
\newcommand{\Goodtype}{\emph{Indicative}}
\newcommand{\badtype}{\emph{encoded}}

\newcommand{\ourattack}{\emph{Single-Node}}
\newcommand{\ouredgeattack}{\emph{Single-Edge}}
\newcommand\para[1]{\vspace{3pt}\noindent \textbf{#1}}
\newcommand\rev[1]{\textcolor{black}{#1}}

\definecolor{ao}{rgb}{0.0, 0.5, 0.0}
\definecolor{mypurple}{HTML}{AB30C4}
\newcommand{\marksize}{1.8pt}

\begin{document}

\begin{frontmatter}

\title{Single-Node Attacks for Fooling Graph Neural Networks}

\author[technion]{Ben Finkelshtein\corref{equal}}
\ead{benfin@campus.technion.ac.il}

\author[technion]{Chaim Baskin \corref{equal} \corref{corr}} %
\ead{chaimbaskin@campus.technion.ac.il}

\author[technion]{Evgenii Zheltonozhskii}
\ead{evgeniizh@campus.technion.ac.il}

\cortext[equal]{Equal contribution}
\cortext[corr]{Corresponding author}

\author[cmu]{Uri Alon}
\ead{ualon@cs.cmu.edu}

\address[technion]{Technion, Haifa, Israel}
\address[cmu]{Carnegie Mellon University}

\begin{abstract}
Graph neural networks (GNNs) have shown broad applicability in a variety of domains.
These domains, e.g., social networks and product recommendations, are fertile ground for malicious users and behavior.
In this paper, we show that GNNs are vulnerable to the extremely limited (and thus quite realistic) scenarios of a single-node adversarial attack, where the perturbed node cannot be chosen by the attacker. 
That is, an attacker can force the GNN to classify any target node to a chosen label, by only slightly perturbing the features or the neighbors list of another single arbitrary node in the graph, even when not being able to select that specific attacker node.
When the adversary is allowed to \emph{select the attacker node}, these attacks are even more effective.
We demonstrate empirically that our attack is effective across various common GNN types (e.g., GCN, GraphSAGE, GAT, GIN) and robustly optimized GNNs (e.g., Robust GCN, SM GCN, GAL, LAT-GCN), outperforming previous attacks across different real-world datasets both in a targeted and non-targeted attacks.
Our code is available anonymously at \url{https://github.com/gnnattack/SINGLE}. 
\end{abstract}

\begin{keyword}
Graph neural networks \sep adversarial robustness \sep node classification  
\end{keyword}

\end{frontmatter}

\section{Introduction}
\label{sec:intro}
\newcommand{\introfigheight}{3.5cm} 

\emph{Graph neural networks} (GNNs) \cite{scarselli2008graph,micheli2009neural} are becoming increasing popular due to their  generality and computation efficiency \cite{kipf2016semi,hamilton2017inductive,velic2018graph,xu2018powerful}. 
Graph-structured data underlie a plethora of domains such as citation networks \cite{sen2008collective}, social networks \cite{ribeiro2017like, ribeiro2018characterizing}, 
knowledge graphs \cite{trivedi2017know,schlichtkrull2018modeling},
and product recommendations \cite{shchur2018pitfalls}. 
Clearly, GNNs are useful in a variety of real-world data. %

\begin{figure*}[t]
        \centering
        \begin{subfigure}{.48\linewidth}
            \centering
            \includegraphics[width=1\linewidth]{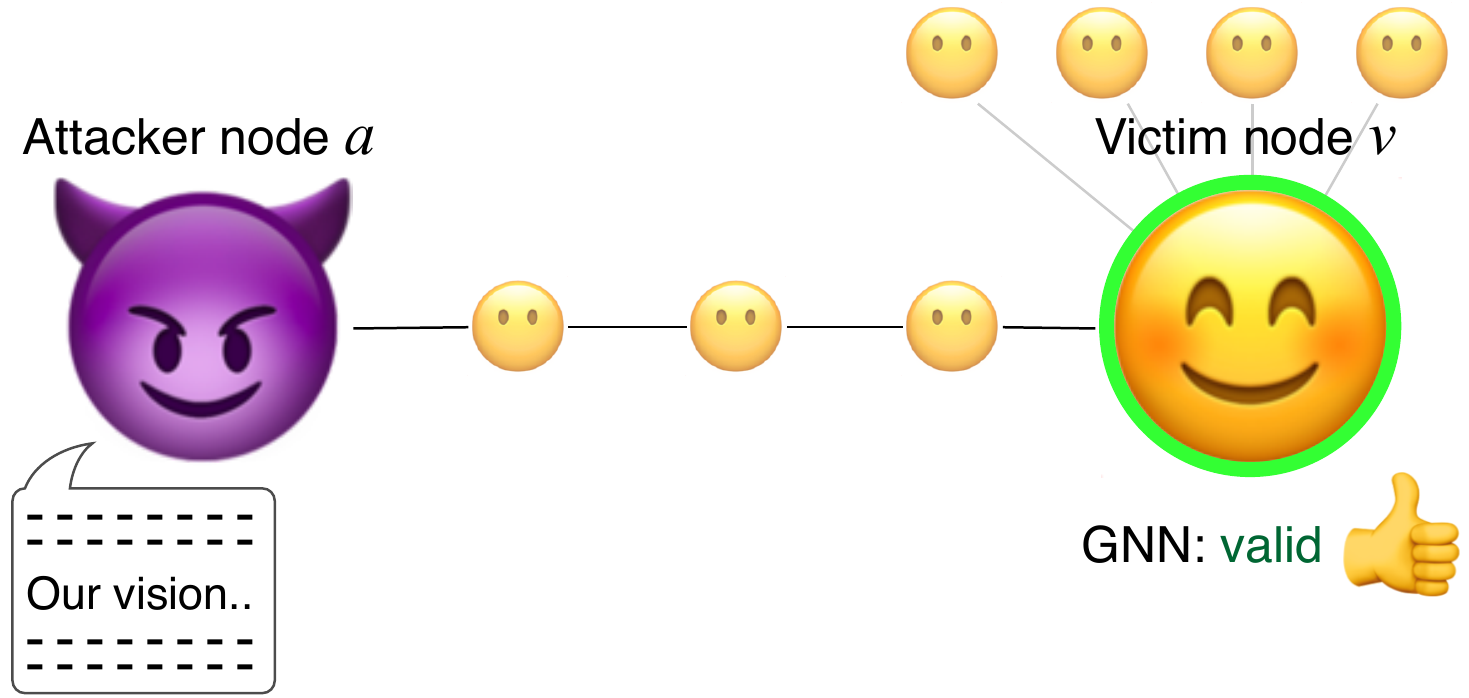}
            \caption{Before attacking: The victim node ($v$) is classified as valid.}
            \label{fig:bottleneck-seq}
        \end{subfigure}
        \hfill
        \begin{subfigure}{.48\linewidth}
            \centering
            \includegraphics[width=1\linewidth]{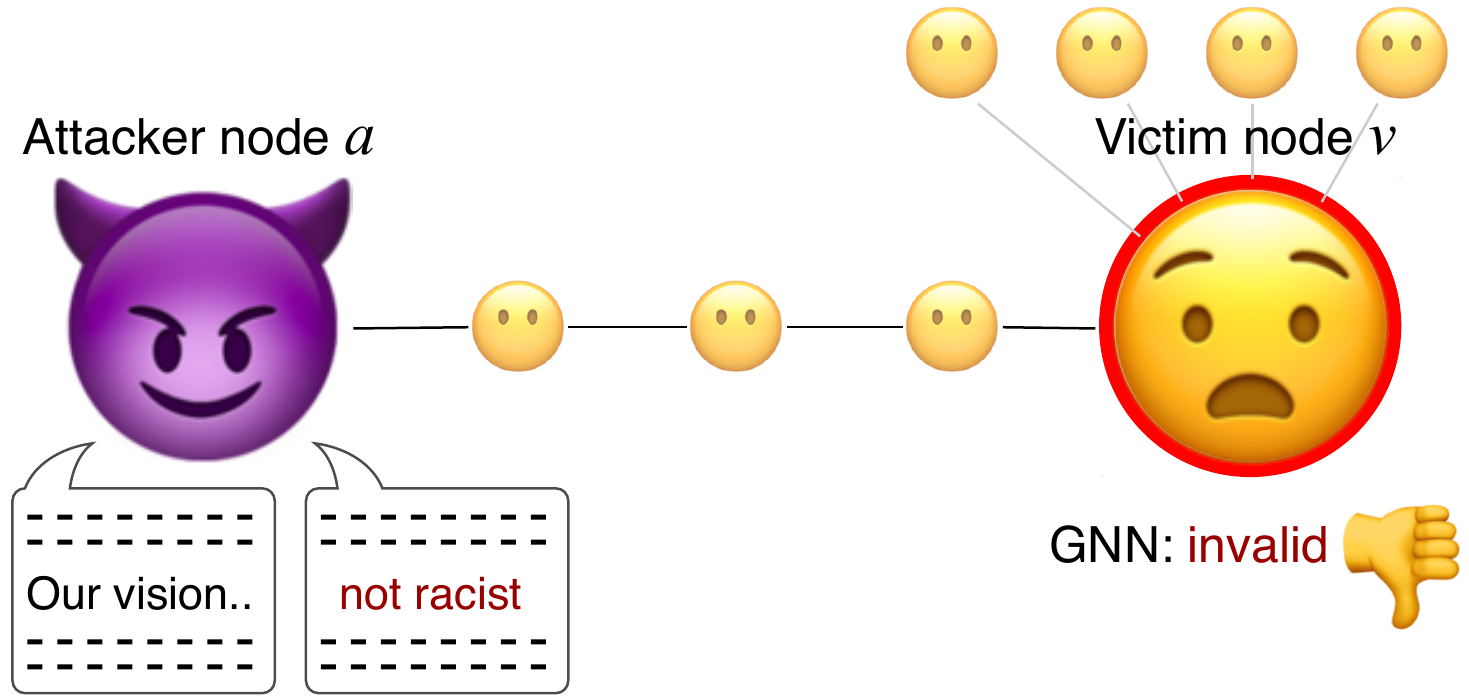}
			\caption{After attacking: The victim node ($v$) is classified as invalid.}
        \end{subfigure} 
        \caption{An partial adversarial example from the test set of the Twitter dataset. %
        An adversarially-crafted post perturbs the representation of the attacker node. This perturbation causes a misclassification of the target victim node, although the two nodes are not even direct neighbors.}
        \label{fig:intro}
\end{figure*}

Most work in this field has focused on designing new GNN variants and applying them to a growing number of domains. 
Very few past works, however, have explored the vulnerability of GNNs to realistic adversarial examples.
Consider the following scenario: a malicious user or a bot joins a social network such as Twitter or Facebook. The malicious user mocks the behavior of a benign user, establishes connections with other users, and submits benign posts. After some time, the user submits a new adversarially crafted post, that might seem irregular but overall appears benign. 
If the social network uses a GNN-based model to detect malicious users,
the new adversarial post changes the representation of the user as seen by the GNN.
As a result, another specific benign user gets blocked from the network; alternatively, another malicious user submits an inciteful or racist post -- but does not get blocked. This scenario is illustrated in \cref{fig:intro}.
In this paper, we show the feasibility of such a troublesome scenario: a single attacker node can perturb its own representation, such that another node will be misclassified as a label of the attacker's choice. 

\rev{Prior work that explored adversarial attacks on GNNs required the perturbation to span \emph{multiple} aspects of the graph: changing features of multiple nodes \cite{zugner2018adversarial}, adding and removing multiple edges \cite{dai2018adversarial,bojchevski2019adversarial,li2020adversarial}, or both \cite{zugner2018adversarial,wu2019adversarial}.
In this paper, we demonstrate the surprising effectiveness of a \emph{single}-node attack.}%

When the attacker node is chosen randomly among nodes within the victim node's reach, 
\rev{our single-node attack reduces test accuracy by (absolute) 7\% on average, across multiple datasets. 
If the attacker node can be \emph{chosen}, for example, by hacking into an existing social network account, the efficiency of the attack significantly increases, and reduces test accuracy by 35\%. }
\rev{We also present a new single-edge adversarial attacks on GNNs,  
where the attacker node is chosen randomly and is limited to either inserting or removing a single edge.
In both single-node and single-edge attacks, we present a white-box gradient-based approach for selecting the attacker. %
Further, we present a black-box, model-free approach that chooses the attacker node using the graph topology.} 
Finally, we perform an extensive experimental evaluation of our approach on multiple datasets and GNN architectures.

To summarize, our contributions are:
\begin{itemize}
  \item We present several single-node adversarial attacks, which perturb a few features of the attacker node. 
  \item We present several single-\emph{edge} attacks that add or remove a single edge in the graph. 
  \item We extend our attacks and allow the attacker node to be chosen according to a black-box model-free approach, or a white-box gradient-based approach, which increases the effectiveness of our attack significantly. 
  \item We show experimentally that our approaches and their variations significantly outperform existing attacks on standard, robust, and adversarially trained GNNs. 
  \item Finally, we perform an extensive ablation study, showing the trade-off between the effectiveness of the attack and its unnoticeability.
\end{itemize}

\section{Related work}
\label{sec:related}
Works on adversarial attacks on GNNs differ in several main aspects. In this section, we discuss the main criteria, to clarify the settings that we address.

\rev{\para{Single vs.\ multiple node  perturbations}
All previous works allowed perturbing \emph{multiple} nodes, or edges that are covered by multiple nodes: 
\citet{zugner2018adversarial} perturb features of a \emph{set} of attacker nodes; \citet{zang2020graph} assume ``a few bad actors''. 
In contrast, we address the extremely limited and more realistic scenario of a \emph{single} attacker node.}

\rev{\para{Edge perturbations}
Most adversarial attacks on GNNs perturb the input graph by allowing the insertion or deletion of \emph{multiple} edges \cite{zugner2018meta,xu2019topology, chen2018fast, li2020adversarial, zugner2018adversarial, chang2020restricted}. Some previous works dealt with the topic of a single-edge attack briefly \cite{bojchevski2019adversarial}. 
\citet{sun2020non} and \cite{dai2018adversarial} tried to guide a reinforcement learning agent to reduce the GNN node classification
performance. \citet{waniek2018hiding} and \citet{chang2020restricted} allowed the insertion and deletion of edges, using attacks that are based on correlations and eigenvalues, and not on gradients.}%
\rev{The attack of \citet{dai2018adversarial} managed to reduce accuracy by only 10\% at most, because it could only \emph{remove} edges.
Unlike previous works, we focus on single-edge attacks, where the choice of the edge is gradient-based and a single attacker is chosen randomly or via a white-box approach.}

\para{Direct vs.\ influence attacks}
Prior works also differ by focusing on either \emph{direct attacks} or \emph{influence attacks}. 
In direct attacks, the attacker \emph{perturbs the victim node itself}. 
For example, the attack of \citet{zugner2018adversarial} is the most effective when the attacker and the target are the same node.
In influence attacks, the perturbed nodes are at least one hop away from the victim node.  In this paper, we show that the unrealistic \emph{direct} assumption is not required (\ourattack{}-\emph{direct} in \cref{subsec:scenario-ablation}), and that our attack is effective \emph{when the attacker and the target are not even direct neighbors}, i.e., they are at least \emph{two} hops away (\ourattack{}-\emph{hops} in Section \cref{subsec:single-node-variants}).

\para{Poisoning vs.\ evasion attacks}
In a related scenario, some work \cite{zugner2018meta, bojchevski2019adversarial, li2020adversarial,zhang2020gnnguard} focused on \emph{poisoning} attacks that perturb examples before training. 
Contrarily, we focus on the standard \emph{evasion} scenario of adversarial examples \cite{szegedy2013intriguing, goodfellow2014explaining}, where the attack operates at test time, after the model is trained. %

\section {Preliminaries}
\label{sec:method}
\rev{Let $\mathcal{G}=\{G_i\}_{i=1}^{N_G}$ be a set of graphs.
Each graph $G=\left(\mathcal{V}, \mX, \mathcal{E}\right)\in \mathcal{G}$ has a set of $N$ nodes $\mathcal{V}$ and a set of edges $\mathcal{E}\subseteq \mathcal{V}\times\mathcal{V}$, where $\left(u,v\right)\in\mathcal{E}$ denotes an edge from a node $u\in\mathcal{V}$ to a node $v\in\mathcal{V}$.
 $\mX\in\sR^{N\times D}$ is a matrix of $D$-dimensional given node features. The $i$-th row of $\mX$ is the feature vector of the node $v_i\in \mathcal{V}$ and is denoted as $\vx_i=\mX_{i,:}\in \sR^D$.}

\para{Graph neural networks.} GNNs operate by iteratively propagating neural messages between neighboring nodes. Every GNN layer updates the representation of every node by combining its current representation with the aggregated current representations of its neighbors. 

Formally, each node is associated with an initial representation $\vx_v^{\left(0\right)}=\vh_v^{\left(0\right)} \in \sR^{D}$. This representation is considered as the given features of the node. Then, a GNN layer updates each node's representation given its neighbors, yielding $\vh_v^{\left(1\right)} \in \sR^{D}$ for every $v\in\mathcal{V}$. In general, the $\ell$-th layer of a GNN is a function that updates a node's representation by combining it with its neighbors: 
\begin{equation*}
	\vh_v^{\left(\ell\right)}=\text{COMBINE}\left(
	\vh_v^{\left(\ell-1\right)}, 
	\{\vh_u^{\left(\ell-1\right)}\mid u\in\mathcal{N}_v\}
	;\theta_{\ell}\right)
	\label{eq:layer}
\end{equation*}
where $\mathcal{N}_v$ is the set of direct neighbors of $v$: $\mathcal{N}_v=\{u \in \mathcal{V} \mid \left( u,v \right) \in \mathcal{E}\}$. 

The COMBINE function is what mostly distinguishes GNN types. For example, graph convolutional networks (GCN) \cite{kipf2016semi} define a layer as:
\begin{equation*}
	\vh_v^{\left(\ell\right)}
	=
	\mathrm{ReLU}\left(
	\sum\nolimits_{u\in \mathcal{N}_v\cup\{v\}}  \frac{1}{c_{u,v}} 
	\mW^{\left(\ell\right)}\vh_{u}^{\left({\ell-1}\right)} 
	\right)
	\label{eq:gcn}
\end{equation*}
where $c_{u,v}$ is a normalization factor usually set to $\sqrt{\left|\mathcal{N}_v \right| \cdot \left|\mathcal{N}_u\right|} $. 
After $\ell$ such aggregation iterations, every node representation captures aggregated information from all nodes within its $\ell$-hop neighborhood. 
The total number of layers $L$ is usually determined empirically as a hyperparameter. The final representation $\vh_v^{\left(L\right)}$ is usually used in predicting properties of the node $v$. 

For brevity, we focus our definitions on the common semi-supervised transductive node classification goal, where the dataset contains a single graph $G$, and the split into training and test sets is across nodes in the same graph.
Nonetheless, these definitions can be trivially generalized to the inductive setting, where the dataset contains multiple graphs, the split into training and test sets is between graphs, and the test nodes are  unseen during training.

We associate each node $ v\in \mathcal{V}$ with a class $y_v \in\mathcal{Y}=\{1,...,Y\}$. %
Labels of training nodes are given during training; test nodes are seen during training but without their labels. %
\rev{Given a training subset $\mathcal{D}=\left(\mX, \mathcal{E}, \{\left(v_i,y_i\right)\}_{i=0}^{N_\mathcal{D}}\right)$,
the goal is to learn a model $f_\theta: \left(\mX, \mathcal{E}, \mathcal{V}\right) \rightarrow \mathcal{Y}$ that will classify the rest of the nodes correctly.
During training, the model $f_\theta$ thus minimizes the loss over the given labels, using $J\left(\cdot,\cdot\right)$, which is typically the cross-entropy loss:
\begin{align}
	\theta^{*}&=\text{argmin}_{\theta} \mathcal{L}\left(f_{\theta},\mathcal{D}\right)= \nonumber\\&= \text{argmin}_{\theta} \frac{1}{N_\mathcal{D}}\sum_{i=0}^{N_\mathcal{D}} J\left(f_{\theta}\left(\mX, \mathcal{E}, v_i\right),y_i\right)
\label{eq:loss}
\end{align}}

\section{Method}
\label{sec:single}
In this section, we describe our \emph{Single-Node indirect gradient adversariaL evasion}  (dubbed \ourattack{}) attack. 
This is the first influence attack that perturbs nodes, which works with an \emph{arbitrary single attacker node} (in contrast to to multiple \cite{zugner2018adversarial} and ``direct'' attacks \cite{li2020adversarial}).
In \cref{sec:single-edge}, we propose a \emph{Single-Edge gradient adversariaL evasion}  (dubbed \ouredgeattack{}) attack.
In \cref{subsec:single-node-variants}, we present a white-box (GradChoice) variation for the selection of the attacker for both \ourattack{} and \ouredgeattack{} along with a black-box (Topology) variation that chooses the attacker node in \ourattack{}, following a heuristic.

\begin{table*}
    \centering
    \begin{tabular}{lrrrr}
        \toprule
        & \multicolumn{1}{c}{\textbf{Cora}} 	& \multicolumn{1}{c}{\textbf{CiteSeer}} & \multicolumn{1}{c}{\textbf{PubMed}}   & \multicolumn{1}{c}{\textbf{Twitter}}\\
        \midrule
		Clean (no attack) & 80.5\eb{0.8} & 68.5\eb{0.7} & 78.5\eb{0.6}  & 89.1\eb{0.2} \\
        \ourattack{} 
        &  71.5\eb{0.8} &  47.7\eb{0.8} &  74.3\eb{0.3}  & 85.5\eb{2.1} \\
        \midrule
        \ourattack{}+\emph{GradChoice} & 64.4\eb{1.0} & 37.5\eb{0.8} & 68.5\eb{0.1}  &  53.9\eb{9.8}\\
        \ourattack{}+\emph{Topology} &   61.2\eb{0.9} &  34.5\eb{1.4} & 66.0\eb{0.5}  & 55.8\eb{11.2}\\
        \midrule
        \ourattack{}-\emph{hops} &  76.5\eb{0.8} & 61.2\eb{0.8} & 75.5\eb{0.2} & 87.8\eb{1.6} \\
        \ourattack{}-\emph{two attackers} & 67.8\eb{0.8} & 42.2\eb{0.6} & 69.9\eb{0.3} &  84.8\eb{3.3}\\
        \ourattack{}-\emph{direct} &  53.4\eb{1.1} &  23.4\eb{1.3} & 46.3\eb{1.1} & 80.1\eb{2.3}\\
        \bottomrule
    \end{tabular}
    \caption{
    Test accuracy (lower is better) under the different variations of a \ourattack{} attack, when the attacker node is chosen \emph{randomly}.}
    \label{tab:single-node-results}
\end{table*} 

\begin{table*}
    \centering
    \resizebox{\linewidth}{!}{
    \begin{tabular}{lrrrr}
        \toprule
        & \multicolumn{1}{c}{\textbf{Cora}} 	& \multicolumn{1}{c}{\textbf{CiteSeer}} & \multicolumn{1}{c}{\textbf{PubMed}}   & \multicolumn{1}{c}{\textbf{Twitter}}\\
        \midrule
		Clean (no attack) & 80.5\eb{0.8} & 68.5\eb{0.7} & 78.5\eb{0.6}  & 89.1\eb{0.2} \\
        \midrule
        $R_{ND}$ \cite{zugner2018adversarial} 
        & 61.0 & 60.0 & - & -\\
        \textsc{Nettack-In} \cite{zugner2018adversarial} 
        & 67.0 & 62.0 & - & -\\
        GF-Attack \cite{chang2020restricted} 
        & 72.6 & 64.7 & 72.4 & -\\
        \midrule
        \ouredgeattack{} (ours) & 70.5\eb{0.6} &  48.2\eb{0.9} &  59.7\eb{0.7} & 82.7\eb{0.1}\\
        \ouredgeattack{}+\emph{GradChoice} (ours) & \textbf{29.7}\eb{2.4} & \textbf{11.9}\eb{0.8} & \textbf{15.3}\eb{0.4}  &  82.0\eb{1.4} \\
        \bottomrule
    \end{tabular}
    }
    \caption{
    Test accuracy (lower is better) under different edge-based attacks.}
    \label{tab:single-edge-results}
\end{table*}

\subsection{Problem Definition}
Given a graph $G=\left(\mathcal{V}, \mX, \mathcal{E}\right)$, a trained model $f_{\theta}$, a ``victim'' node $v$ from the test set along with its classification by the model $\hat{y}_{v}=f_{\theta}\left(v, \mX, \mathcal{E}\right)$, we assume that an adversary controls another node $a$ in the graph. 
The goal of the adversary is to modify its own feature vector $\vx_a$ by adding a perturbation vector $\noise\in \sR^{D}$ of its choice, such that the model's classification of $v$ will change. 
We denote by $\mX_{a}^{\noise}$
the node feature matrix, where vector $\noise$ was added with to row of $\mX$ that corresponds to  node $a$.
In a non-targeted attack, the goal of the attacker is to find a perturbation vector $\noise$ that will change the classification to \emph{any} other class, i.e., $f_{\theta}\left(\mathcal{E}, \mX_{a}^{\noise}, v\right) \neq f_{\theta}\left(\mathcal{E}, \mX, v\right)$.
In a \emph{targeted} attack, the adversary chooses a specific label $y_{adv} \in \mathcal{Y}$ and the adversary's goal is to force $f_{\theta}\left(\mathcal{E}, \mX_{a}^{\noise}, v\right) = y_{adv}$.

In this work, we focus on gradient-based attacks. These attacks assume that the attacker can access a similar model to the model under attack and compute gradients.
As recently shown by \citet{wallace2020stealing}, this is a reasonable assumption: an attacker can query the original model, imitate the model under attack by training an imitation model, find adversarial examples using the imitation model, and transfer these adversarial examples back to the original model. Under this assumption, these attacks are general and are applicable to any GNN and dataset.

\subsection{Limited Perturbations}
\label{subsec:limit}
Our main challenge is to find a {\bf realistic} adversarial perturbation in a way that allows us to constrain its magnitude.
In images, this is usually attained by constraining the
$l_{\infty}$-norm of the perturbation vector $\noise$. It is, however, unclear how one can constrain a graph.

In most GNN datasets, a node's features are a bag-of-words representation of the words that are associated with the node. For example, in Cora \cite{mccallum2000automating,sen2008collective}, every node is annotated by a many-hot feature vector of words that appear in the paper. We denote such datasets as \emph{discrete datasets}, because the given feature vector of every node contains only discrete values. In contrast, in PubMed \cite{Namata2012QuerydrivenAS}, node vectors are TF-IDF word frequencies; in 
Twitter-Hateful-Users \cite{ribeiro2017like}, node features are averages of GloVe embeddings, which can be viewed as word frequency vectors multiplied by a (frozen) embedding matrix.
We denote such datasets as \emph{continuous datasets}, because the initial feature vector of every node is continuous. 

In continuous datasets, the number of times a word has been added or removed by the perturbation should not seem anomalous. Therefore, we constrain the perturbation vector $\noise$ by requiring $\norm{\noise}_{\infty} \leqslant \epsilon_{\infty}$ -- the absolute value of the elements in the perturbation vector is bounded by $\epsilon_{\infty} \in \sR^{+}$. However, imagine a random post on twitter, solely constraining the $l_{\infty}$-norm of our perturbation vector. As the average post uses a small set of words from the corpus, our perturbed post, which make use of a larger part of the corpus (with a low frequency) would be easily detected.

To prevent easy detection of the attack, we want the perturbed sample to be in the domain of the training data distribution, preferably with a high likelihood. At the very least, for any choice of norm function, the norm of the perturbed sample features should be comparable with the norm of training samples features.

We limit the number of words (features) we perturb in a node $\norm{\noise}_0$ to be strictly smaller than the average number of non-zero entries in the dataset. In this way, the number of words added or removed by the perturbation would not seem anomalous and our perturbation would be indiscernible.

In summary, for continuous datasets, we limit the value of the features we perturb to be strictly smaller than the average over the value of the non-zero entries in the dataset $\norm{\noise}_{\infty} \leqslant \epsilon_{\infty}$ while also requiring that $\nicefrac{\norm{\noise}_{0}}{D} \leqslant \epsilon_{0}$: the fraction of non-zero elements in the perturbation vector is bounded by $\epsilon_{0} \in [0,1]$.

For discrete datasets the perturbed vector $\vx_a\mathsmaller{+}\noise$ must be discrete as well -- if every node is given as a many-hot vector $\vx_a$, the perturbed vector $\vx_a\mathsmaller{+}\noise$ must remain many-hot as well. Hence, we constrain the $l_{0}$-norm of our perturbation vector. Where for discrete datasets, measuring the $\ell_0$ norm of $\noise$ is equivalent to  the $\ell_1$ norm: $\norm{\noise}_{0}=\norm{\noise}_{1}$.

\subsection{Finding the Perturbation Vector}
\label{subsec:finding}
To find the perturbation vector, our general approach is to iteratively differentiate the desired loss of $v$ with respect to the perturbation vector $\noise$, and update $\noise$ according to the gradient, similarly to the general approach in adversarial examples of computer vision models \cite{goodfellow2014explaining}.
In non-targeted attacks, we take the positive gradient of the loss of the undesired label to increase the loss;
in targeted attacks, we take the negative gradient of the loss of the adversarial label $y_{adv}$:
\begin{gather*}
	\noise_{t+1}=
	\begin{cases} 
	\noise_{t} + \gamma \nabla_{\noise}J\left(f_{\theta}\left(\mX_{a}^{\noise_{t}}, \mathcal{E}, v\right), \hat{y}_v\right) & \text{non-targeted } \\
	\noise_{t} - \gamma \nabla_{\noise}J\left(f_{\theta}\left(\mX_{a}^{\noise_{t}}, \mathcal{E}, v\right), y_{adv}\right) & \text{targeted }	
	\end{cases}
	\label{eq:perturbation}
\end{gather*}
where $\gamma \in \sR^{+}$ is a learning rate. We repeat this process for a predefined number of $K$ iterations, or until the model predicts the desired label.

\para{In continuous datasets}, %
after each update, we
clip perturbation vector $\noise_{t+1}$ according to the $\epsilon_\infty$ constraint: $\norm{\noise_{t+1}}_{\infty} \leqslant \epsilon_{\infty}$ and set all attributes of perturbation vector $\noise_{t+1}$ to zero, except for the largest $\epsilon_0$$\cdot$$D$ attributes, according to the $\ell_0$ constraint:
$\nicefrac{\norm{\noise_{t+1}}_{0}}{D} \leqslant \epsilon_{0}$.

\para{In discrete datasets}, %
where node features are many-hot vectors, the only possible perturbation to every feature is ``flipping'' it from $0$ to $1$ or vice versa.
In every update iteration, we thus ``flip'' the vector attribute with the largest gradient out of the vector attributes that have not been yet flipped.
We repeat this process as long as the $\ell_0$ constraint holds: $\nicefrac{\norm{\noise_{t+1}}_{0}}{D} \leqslant \epsilon_{0}$,
or until the model predicts the desired label.

\para{Differentiate by frequencies, not by embeddings.} 
When taking the gradient %
$\nabla_{\noise}$, there is a subtle, but crucial, difference between the way that node representations are provided in the dataset:
\begin{inparaenum}[(a)]
\item \goodtype{} datasets provide initial node representations $\mX = \left[\vx_1, \vx_2, ...\right]$ that are word indicator vectors (many-hot) or frequencies such as (weighted) bag-of-words \cite{sen2008collective, shchur2018pitfalls};
\item in contrast, in \badtype{} datasets, initial node representations are already encoded, e.g., as an average of word2vec vectors \cite{hamilton2017inductive}.  
\end{inparaenum}
\Goodtype{} datasets can be converted to \badtype{} by multiplying every vector by an embedding matrix; \badtype{} datasets \emph{cannot} be converted to \goodtype{}, without the authors releasing the textual data that was used to create the \badtype{} dataset.

In \goodtype{} datasets, a perturbation of a node vector \emph{can} be realized as a perturbation of the original text from which the \goodtype{} vector was derived. That is, adding or removing words in the text can result in the perturbed node vector.
In contrast, a few-indices perturbation in \badtype{} datasets might be an effective attack, but is \emph{not} be realistic because there is no perturbation of the original text that results in that perturbation of the vector. 
In other words, realistic adversarial examples require \goodtype{} datasets, or converting \badtype{} datasets to \goodtype{} representation %
(as we do in \cref{sec:exp}) using their original text.

\subsection{Single-Edge GNN Attack}
\label{sec:single-edge}
\ouredgeattack{} is an attack that either inserts or removes a single edge according to the gradient,\footnote{This can be  implemented easily using \emph{edge weights}: training the GNN with weights of $1$ for existing edges, adding all possible edges with weights of $0$, and taking the gradient with respect to the vector of weights.} of an \emph{arbitrary single attacker node}. We denote the insertion or removal of an edge as a toggle operator that ``flips'' the current status of the edge.

\para{Finding an edge to flip}
For each node under attack $v$ and a corresponding attacker node $u$, we add to the original topology of the graph $\mathcal{E}$  edges between the attacker node $u$ and every node in the (k-1)-hop vicinity of the attacked node $\mathcal{N}_{k-1}\left(v\right)$, as nodes that are further away would not influence the attacked node:
\begin{align}
   \mathcal{E}\left(v\right) = \qty{(u,w)| \forall w\in\mathcal{V} \text{ s.t. } (u,w)\in\mathcal{E} \cup w\in \mathcal{N}_{k-1}\left(v\right) } 
\end{align}

We also add an edge weight vector $W\left(v\right)$. To preserve the original topology we initialize the weight of the existing edges in $\mathcal{E}$ to 1 and all other edges to 0.
We then choose the edge $e\left(v\right) \in \mathcal{E}\left(v\right)$ with the largest loss gradient and flip it:
\begin{gather*}
	e^{*}\left(v\right) = \mathrm{argmax}_{e\left(v\right)}
	\begin{cases} 
	-J\left(f_{\theta}\left(X, \mathcal{E}\left(v\right), W\left(v\right), v\right), \hat{y}_v\right) & \text{non-targeted } \\
	J\left(f_{\theta}\left(X, \mathcal{E}\left(v\right), W\left(v\right), v\right), y_{adv}\right) & \text{targeted }	
	\end{cases}
	\label{eq:perturbation2}
\end{gather*}

Similarly to our single-node approach, for non-targeted attacks we take the negative gradient of the loss of the true label to decrease the loss;
for targeted attacks, we take the positive gradient of the loss of the adversarial label $y_{adv}$.

\subsection{Attacker Choice}
\label{subsec:single-node-variants}
If the attacker could \emph{choose} the node it will use for the attack, e.g., by hijacking a specifically chosen existing account in a social network, could they increase the effectiveness of the attack? We examine the following approaches of choosing the attacker node.

\textbf{\emph{\ourattack{} Gradient Attacker Choice} (\ourattack{}+\emph{GradChoice})} chooses the attacker node according to the largest gradient with respect to the node representations (for a non-targeted attack): $a^{*}=\text{argmax}_{a_i\in \mathcal{V}} \norm{\nabla_{\vx_i}J\left(f_{\theta}\left(G,v\right),\hat{y}_v\right)}_{\infty}$. The chosen attacker node is never the victim node itself.

\textbf{\emph{\ourattack{} Topological Attacker Choice} (\ourattack{}+\emph{Topology})} chooses the attacker node according to the graph's topological properties. As an example, we choose the neighbor of the victim node $v$ with the smallest number of neighbors as we expect a node with less neighbor to have more influence over the small amount of neighbors he does have (same intuition stands behind GNN normalization): $a^{*}=\text{argmin}_{a\in \mathcal{N}_v} \left|\mathcal{N}_{a}\right|$.
In this approach, the attacker choice is \emph{model-free}: if the attacker cannot compute gradients, they can at least choose the most harmful attacker node, and then perform the perturbation itself using other non-gradient approaches \cite{waniek2018hiding,chang2020restricted}.

\textbf{\emph{Edge Gradient Attacker Choice (GradChoice)}} is a modification where the edge that is either inserted or removed is sampled from the entire graph, according to the gradient. We compare our two \ouredgeattack{} approaches with additional approaches from the literature. As in \ourattack{}, we report the means and standard deviations. 

\section{Evaluation}
\label{sec:exp}
\rev{In this section, we evaluate the effectiveness of our \ourattack{} and \ouredgeattack{} attacks.  %
In \cref{subsec:results}, we demonstrate the effectiveness of our limited perturbation \ourattack{} attack, including its white-box (GradChoice) and black-box (Topology) variations.
We also show the effectiveness of our \ouredgeattack{} attack, which is higher than some multi-edge attacks. %
In \cref{subsec:adversarial-training} we show that \ourattack{} is robust to adversarial training and robust GNNs (e.g., Robust GCN, SM GCN, GAL, LAT-GCN)}. 
In \cref{subsec:sensitivity,subsec:limit-l0} we analyze the effects of $\epsilon_0$ and $\epsilon_\infty$ , respectively, and in the supplementary material we examine their trade-off.

\para{Setup.} 
We trained \rev{each standard GNN type} 
with two layers ($L=2$), using the Adam optimizer, early stopped according to the validation set, and applied a dropout of $0.5$ between layers. \rev{We trained each robust GNN according to the author' implementation.}
We used up to $K=20$ attack iterations. %
All experiments described in \cref{sec:exp} were performed with GCN, except for \cref{subsec:sensitivity}, where additional GNN types (GraphSAGE \cite{hamilton2017inductive}, GAT \cite{velic2018graph}, and GIN \cite{xu2018powerful}) are used.
In the supplementary material, we show consistent results across all GNN types mentioned above as well as  SGC \cite{wu2019simplifying}. %
We ran each approach five times with different random seeds for each dataset, and report the means and standard deviations.
Our PyTorch Geometric \cite{fey2019pytorchgeometric} implementation is available anonymously at \url{https://github.com/gnnattack/SINGLE}.

\para{Data.}
We used Cora and CiteSeer \cite{sen2008collective}, which are discrete datasets, i.e., the given node vectors are many-hot vectors. 
We also used PubMed \cite{sen2008collective} and the Twitter-Hateful-Users (or, shortly, Twitter) \cite{ribeiro2017like}  datasets, which are continuous, and node features represent frequencies of words.
\rev{As explained in \cref{subsec:limit}, we limit the fraction of perturbed attributes $\epsilon_0$ for all datasets and the absolute change in each element $\epsilon_{\infty}$ for continuous datasets, which allows finer control over the intensity of the perturbation. The $\epsilon_0$ and $\epsilon_\infty$ values for each dataset are provided in the supplementary material. In practice, the attack usually uses \emph{fewer} node attributes. An analysis of values of $\epsilon_0$ and $\epsilon_{\infty}$ is presented in \cref{subsec:sensitivity,subsec:limit-l0}, and in supplementary material.}

The Twitter dataset is originally provided as an \badtype{} dataset, where every node is an average of GloVe vectors \cite{pennington2014glove}. We reconstructed this dataset using the original text from \citet{ribeiro2017like}, to be able to compute gradients with respect to the weighted histogram of words rather than the embeddings. We took the most frequent 10,000 words as node features and used GloVe-Twitter embeddings to multiply by the node features. We thus converted this dataset to indicative rather than encoded.
Statistics of all datasets are provided in the supplementary material.

\begin{table*}
    \centering
\resizebox{\linewidth}{!}{    
    \begin{tabular}{lrrrrr}
        \toprule 
        & \multicolumn{2}{c}{Robust GCN} & \multicolumn{3}{c}{SM GCN}\\
        & \textbf{Cora} & \textbf{CiteSeer} & \textbf{Cora} & \textbf{CiteSeer} & \textbf{PubMed}\\
		\midrule
		Clean & 79.7\eb{0.8} & 58.0\eb{1.9} & 78.8\eb{0.3} & 68.2\eb{0.5} & 78.2\eb{0.6}\\
        \ourattack{} & 74.4\eb{0.7} & 44.5\eb{0.5} & 48.8\eb{1.4} & 22.1\eb{1.6} & 65.7\eb{0.4} \\
        \midrule
        \ourattack{}+\emph{GradChoice} & 69.5\eb{0.5} & 33.8\eb{0.7} & 42.3\eb{1.0} & 18.9\eb{0.3} & 63.7\eb{0.3}\\
        \ourattack{}+\emph{Topology} & 66.5\eb{0.8} & 29.5\eb{1.1} & 38.7\eb{1.0} & 16.4\eb{0.7} & 62.3\eb{0.3}\\
        \midrule
        \ourattack{}-\emph{hops} & 79.4\eb{0.8} & 56.8\eb{1.9} & 52.4\eb{1.6} & 29.3\eb{0.6} & 65.9\eb{0.7}\\
        \ourattack{}-\emph{two attackers} & 69.8\eb{1.0} &  37.2\eb{1.5} & 45.4\eb{1,3} & 21.8\eb{0.8} & 64.2\eb{0.3}\\
        \ourattack{}-\emph{direct} & 54.2\eb{1.3} & 18.6\eb{2.8} & 32.4\eb{0.4} & 13.8\eb{0.6} & 29.8\eb{0.5}\\
        \bottomrule
    \end{tabular}
    }
    \caption{
     Test accuracy of a robustly trained GCN model \cite{zugner2019certifiable} and a GCN with a Soft Medoid as the aggregation function \cite{geisler2020reliable}}
    \label{tab:robust-untargeted-1}
\end{table*}

\subsection{Main Results}
\label{subsec:results}
\cref{tab:single-node-results} presents our main results for non-targeted attacks.  %
Even under the heavy limitations of a single node perturbation, which is limited by the $\ell_0$ and the $\ell_\infty$ norms, \ourattack{} is effective across all datasets.

\ourattack{}-\emph{hops},  which is more indiscernible than attacking with a neighboring node, reduces test accuracy by an average of only 5\% (absolute), whereas \ourattack{}, which attacks using either a neighboring or non-neighboring node, reduces test accuracy by an average of 11\% (absolute).

Choosing the attacker node, whether by using gradients (white-box attack) or topology (black-box), significantly increases the effectiveness of our \ourattack{} attack: for example, in Cora, from 80.5\% (\cref{tab:single-node-results}) to 71.5\% test accuracy. Furthermore, \ourattack+\emph{Topology} outperforms \ourattack+\emph{GradChoice} across all of the datasets, apart from the Twitter dataset. We believe that white box attack is less effective due to the iterative nature of the attack: it is difficult to  differentiate between multiple harmful attackers based on a single gradient step.

\cref{tab:single-edge-results} shows our results for \ouredgeattack{} non-targeted attacks compared to the previous \emph{multiple}-edge attacks: $R_{ND}$ \cite{zugner2018adversarial}, \textsc{Nettack-In} \cite{zugner2018adversarial} and GF-Attack \cite{chang2020restricted} .  %
\ouredgeattack{} is more effective than previous methods over CiteSeer and PubMed, reducing the test accuracy by 20.3\% and 18.8\%, respectively. 
Furthermore, allowing the attacker to perturb an edge from the entire graph using \ouredgeattack+\emph{GradChoice} significantly increases the effectiveness of our \ouredgeattack{} attack. 
As a result, \ouredgeattack+\emph{GradChoice} is the most effective edge-based attack across all datasets: for example, in PubMed, test accuracy drops from 59.7\% (\cref{tab:single-edge-results}) to 15.3\%.
In the supplementary material, we show that allowing \ouredgeattack{}+\emph{GradChoice} to insert and remove \emph{multiple} edges of the same attacker node  does not lead to a significant improvement. %

\subsection{Effectiveness of the Attack Facing Defense}
\label{subsec:adversarial-training}
In this section, we investigate to what extent  defensive training approaches can  defend against \ourattack{}.

\para{Adversarial Training.} We experimented with attacking models that were adversarially trained \cite{madry2018towards}.
In each training step we used \ourattack{} or \ourattack{}+\emph{Topology} on each labeled training node. The model is then trained to minimize the original cross-entropy loss and the adversarial loss: 
\begin{align}
\mathcal{L}\qty(f_{\theta},\mathcal{D}) = \frac{1}{2N_\mathcal{D}}\sum_{i=0}^{N_\mathcal{D}} 
\bigg[ & J\left(f_{\theta}\left(\mX, \mathcal{E},v_i\right),y_i\right)
+ 
J\left(f_{\theta}\left(\mX_{a_i}^{\noise_{i}}, \mathcal{E},v_i\right),y_i\right)\bigg] 
\end{align}
The main difference from \cref{eq:loss} is the adversarial term $J\left(f_{\theta}\left(\mX_{a_i}^{\noise_{i}}, \mathcal{E},v_i\right),y_i\right)$, where $a_i$ is the randomly sampled attacker for node $v_i$ in \ourattack{} or the node chosen according to the topological properties of the graph in \ourattack{}+\emph{Topology}.

After the model is trained, we attack the model with the different variations of a \ourattack{} attack. This is similar to the approach of \citet{feng2019graph} and \citet{deng2019batch}. Instead of using  adversarial training as a  regularization to improve the accuracy of a model on clean data,  here we use adversarial training to defend a model against an attack at test time.

As shown in the supplementary material, adversarial training does not improve the model robustness against the different \ourattack{} attacks. Since, the adversarially trained model is much less susceptible to adversarial attacks, and due to the small size of training set, it appears that adversarially trained models are not able to generalize to unseen nodes. 

\para{Robust Training.} We also experimented with attacking robust models such as robustly trained GCN \cite{zugner2019certifiable},  where the architecture and the training scheme of GNNs are optimized for robustness, Soft Medoid GCN \cite{geisler2020reliable}, which employs a robust aggregation function, GAL \cite{liao2021information}, which locally filters out pre-determined sensitive attributes via adversarial training with the total variation and the Wasserstein distance and LAT-GCN \cite{jin2020robust}, which perturbs the latent representation of a GNN. We used the publicly available code of \citet{zugner2019certifiable}, \citet{geisler2020reliable} and \citet{liao2021information}. For LAT-GCN \cite{liao2021information}, we used our re-implementation. 
\cref{tab:robust-untargeted-1} shows that robust GNNs are as vulnerable to our \ourattack{} attack as a standard GCN, demonstrating the effectiveness of our attack and indicating that there is still much room for novel ideas and improvements to the robustness of current GNNs. Additional results for robust GNN architectures are presented in the supplementary material.

\subsection{Scenario Ablation}
\label{subsec:scenario-ablation}
The main scenario that we focus on in this paper is a \ourattack{} approach that always perturbs a \emph{single} node, which is not the victim node ($a\neq v$). For each victim node, the attacker node is selected randomly and the attack perturbs the chosen attacker node's features according to the $\epsilon_{\infty}$ and $\epsilon_{0}$ values.
We now examine our \ourattack{} attack in other, easier but less realistic, scenarios:

\textbf{\emph{\ourattack{}-hops}} is a modification of \ourattack{} where the attacker node is sampled \emph{only among nodes that are not direct neighbors}, i.e., the attacker and the victim are not directly connected ($\left(a,v\right) \notin \mathcal{E}$). The idea in \emph{\ourattack{}-hops} is to evaluate a  variant of \emph{\ourattack{}} that is more indiscernible in reality.

\textbf{\ourattack{}-\emph{two attackers}} follows \citet{zugner2018adversarial} and \citet{zang2020graph}. It randomly samples two attacker nodes and perturbs their features using the same approach as \ourattack{}. 

\textbf{\ourattack{}-\emph{direct}} perturbs the victim node directly (i.e., $a=v$), an approach that was found to be the most efficient by \citet{zugner2018adversarial}. \cref{tab:single-node-results} shows the test accuracy of these ablations.
Expectantly, perturbing two attacker nodes or perturbing the victim node itself is more effective, albeit less realistic.

\paragraph{Larger number of attackers}
We performed additional experiments with up to five randomly sampled attacker nodes simultaneously and perturb their features using the same approach as \ourattack{}, presented in supplementary material.

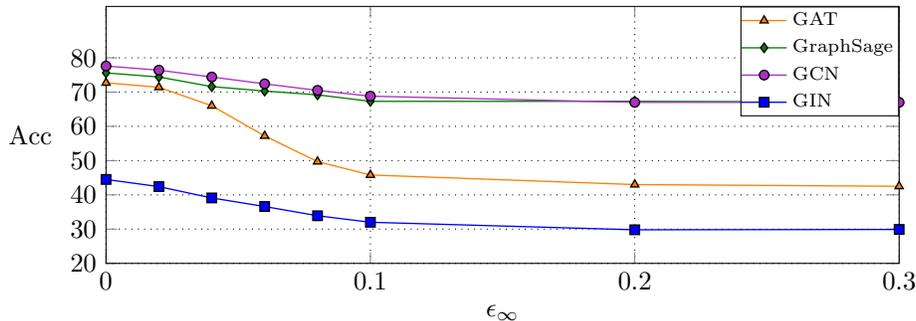
\begin{figure}
\centering
\begin{tikzpicture}[scale=1]
	\begin{axis}[
		xlabel={$\epsilon_\infty$},
		ylabel={Acc},
		x label style={at={(axis description cs:0.5,0.03)},anchor=north},
		y label style={at={(axis description cs:0.25,0.5)},anchor=east},
		ylabel near ticks,
        legend style={at={(1, 1)},anchor=north east,font=\scriptsize,inner xsep=0pt, inner ysep=0pt}, 
        legend cell align={left},
        xmin=0, xmax=0.3,
        ymin=20.0, ymax=95.0,
        xtick={0,0.1,0.2,0.3,0.4,0.5,0.6,0.7,0.8,0.9,1},
        ytick={10,20,...,80},
        ylabel style={rotate=-90}, %
        grid = major,
        major grid style={dotted,black},
        height = 5.0cm,
        width = 1\linewidth,
    ]

\addplot[color=orange, mark options={solid, fill=orange, draw=black}, line width=0.5pt, mark size=\marksize, mark=triangle*, visualization depends on=\thisrow{alignment} \as \alignment, nodes near coords, point meta=explicit symbolic,
    every node near coord/.style={anchor=\alignment, font=\scriptsize}] 
table [meta index=2]  {
x   y       label   alignment
0 72.7		{}		-129
0.02 71.4		{}		-129
0.04 66.0		{}		-129
0.06 57.2		{}		-129
0.08 49.7		{}		-129
0.1	45.8		{}		-129
0.2	43.0		{}		-129
0.3	42.5		{}		-129
	};
    \addlegendentry{GAT}

\addplot[color=ao, mark options={solid, fill=ao, draw=black}, line width=0.5pt, mark size=\marksize, mark=diamond*, visualization depends on=\thisrow{alignment} \as \alignment, nodes near coords, point meta=explicit symbolic,
    every node near coord/.style={anchor=\alignment, font=\scriptsize}] 
table [meta index=2]  {
x   y       label   alignment
0 75.6		{}		-129
0.02 74.4		{}		-129
0.04 71.6		{}		-129
0.06 70.3		{}		-129
0.08 69.2		{}		-129
0.1	67.3		{}		-129
0.2	67.3		{}		-129
0.3	67.1		{}		-129
	};
    \addlegendentry{GraphSage}

\addplot[color=mypurple, mark options={solid, fill=mypurple, draw=black}, line width=0.5pt, mark size=\marksize, mark=*,  visualization depends on=\thisrow{alignment} \as \alignment, nodes near coords, point meta=explicit symbolic,
    every node near coord/.style={anchor=\alignment, font=\scriptsize}] 
table [meta index=2]  {
x   y       label   alignment
0 77.6		{}		-129
0.02 76.4		{}		-129
0.04 74.4		{}		-129
0.06 72.4		{}		-129
0.08 70.5		{}		-129
0.1	68.8		{}		-129
0.2	67.0		{}		-129
0.3	67.0		{}		-129
	};
    \addlegendentry{GCN}

\addplot[color=blue, mark options={solid, fill=blue, draw=black}, line width=0.5pt, mark size=\marksize,  mark=square*, visualization depends on=\thisrow{alignment} \as \alignment, nodes near coords, point meta=explicit symbolic,
    every node near coord/.style={anchor=\alignment, font=\scriptsize}] 
table [meta index=2]  {
x   y       label   alignment
0 44.5		{}		-129
0.02 42.4		{}		-129
0.04 39.1		{}		-129
0.06 36.6		{}		-129
0.08 33.9		{}		-129
0.1	    32.0		{}		-129
0.2	    29.8		{}		-129
0.3	    29.9		{}		-129
	};
    \addlegendentry{GIN}

	\end{axis}
\end{tikzpicture}
\caption{Effectiveness of our \ourattack{} attack compared to the allowed  $\epsilon_{\infty}$  (on PubMed).
}
\label{fig:epsilons}
\end{figure}

\subsection{Sensitivity to $\epsilon_{\infty}$}
\label{subsec:sensitivity}

How does the norm of the adversarial perturbation affect the attack? Intuitively, the less we restrict the perturbation (i.e., the larger  the value of $\epsilon_{\infty}$ is), the more powerful the attack. We examine whether this holds in practice.

In our experiments described in \cref{subsec:results,subsec:adversarial-training,subsec:scenario-ablation}, we used $\epsilon_{\infty}=0.1$ for the continuous datasets (PubMed and Twitter).
Here, we vary the value of $\epsilon_{\infty}$ across different GNN types. %
\cref{fig:epsilons} shows the results on PubMed and demonstrates that smaller values of $\epsilon_{\infty}$ are effective as well. 
As the value of $\epsilon_{\infty}$ increases, GAT \cite{velic2018graph} demonstrates a large drop in test accuracy. In contrast, GCN, GraphSage and and GIN \cite{xu2018powerful} are more robust to an increased norm of perturbations.

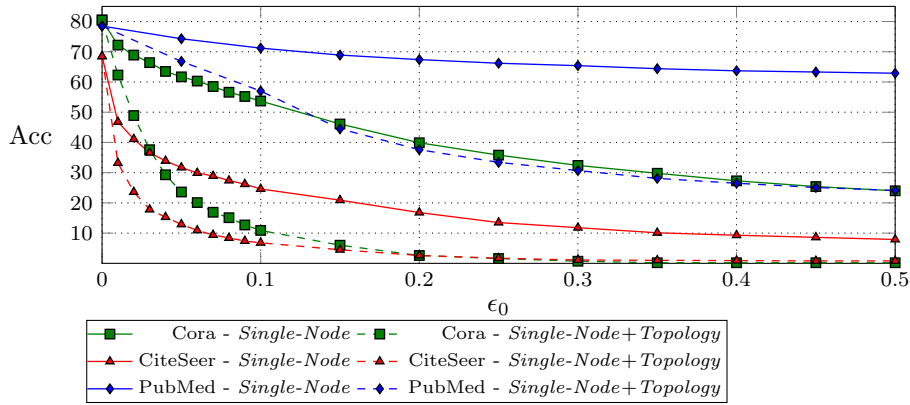
\begin{figure}
\centering
	\begin{tikzpicture}[scale=1]
	\begin{axis}[
		xlabel={$\epsilon_{0}$},
		ylabel={Acc},
		x label style={at={(axis description cs:0.5,0.05)},anchor=north},
		y label style={at={(axis description cs:0.25,0.5)},anchor=east},
		ylabel near ticks,
        legend  style={at={(-0.02,-0.22)},anchor=north west,
         legend columns=2 ,
        font=\scriptsize, inner xsep=0pt, inner ysep=0pt},
        legend cell align={right},
        xmin=0.0, xmax=0.5,
        ymin=0.0, ymax=85.0,
        xtick={0,0.10,0.20,...,1},
        ytick={10,20,...,100},
        ylabel style={rotate=-90}, %
        tick label style={font=\footnotesize} ,
        grid = major,
        major grid style={dotted,black},
        height = 5.0cm,
        width = 1\linewidth,
    ]
    
    \addplot[color=ao, mark options={solid, fill=ao, draw=black}, line width=0.5pt, mark size=\marksize, mark=square*, visualization depends on=\thisrow{alignment} \as \alignment, nodes near coords, point meta=explicit symbolic,
    every node near coord/.style={anchor=\alignment, font=\scriptsize}] 
table [meta index=2]  {
x   y       label   alignment
0.00	80.5		{}		-129
0.01	72.2		{}		-129
0.02	68.9		{}		-129
0.03	66.4		{}		-129
0.04	63.5		{}		-129
0.05	61.7		{}		-129
0.06	60.3		{}		-129
0.07	58.5		{}		-129
0.08	56.6		{}		-129
0.09	55.2		{}		-129
0.10	53.7		{}		-129
0.15	46.1		{}		-129
0.20	39.9		{}		-129
0.25	35.8		{}		-129
0.30	32.4		{}		-129
0.35	29.8		{}		-129
0.40	27.3		{}		-129
0.45	25.4		{}		-129
0.50	24.0		{}		-129
0.55	22.1		{}		-129
0.60	21.4		{}		-129
0.65	20.7		{}		-129
0.70	20.6		{}		-129
0.75	20.6		{}		-129
0.80	20.6		{}		-129
0.85	20.6		{}		-129
0.90	20.6		{}		-129
0.95	20.6		{}		-129
1.00	20.6		{}		-129
	};
    \addlegendentry{Cora - \ourattack{}}
    
    \addplot[dashed, color=ao, mark options={solid, fill=ao, draw=black}, line width=0.5pt, mark size=\marksize, mark=square*, visualization depends on=\thisrow{alignment} \as \alignment, nodes near coords, point meta=explicit symbolic,
    every node near coord/.style={anchor=\alignment, font=\scriptsize}] 
table [meta index=2]  {
x   y       label   alignment
0.00	80.5		{}		-129
0.01	62.3		{}		-129
0.02	48.9		{}		-129
0.03	37.6		{}		-129
0.04	29.3		{}		-129
0.05	23.6		{}		-129
0.06	20.1		{}		-129
0.07	16.9		{}		-129
0.08	15.1		{}		-129
0.09	12.7		{}		-129
0.10	10.9		{}		-129
0.15	6.0		{}		-129
0.20	2.6		{}		-129
0.25	1.6		{}		-129
0.30	0.7		{}		-129
0.35	0.2		{}		-129
0.40	0.2		{}		-129
0.45	0.2		{}		-129
0.50	0.2		{}		-129

	};
	\addlegendentry{Cora - \ourattack{}+\emph{Topology}}
	    
    \addplot[color=red, mark options={solid, fill=red, draw=black}, line width=0.5pt, mark size=\marksize, mark=triangle*, visualization depends on=\thisrow{alignment} \as \alignment, nodes near coords, point meta=explicit symbolic,
    every node near coord/.style={anchor=\alignment, font=\scriptsize}] 
table [meta index=2]  {
x   y       label   alignment
0.00	68.5		{}		-129
0.01	46.8		{}		-129
0.02	41.1		{}		-129
0.03	36.6		{}		-129
0.04	33.9		{}		-129
0.05	31.7		{}		-129
0.06	29.9		{}		-129
0.07	28.9		{}		-129
0.08	27.4		{}		-129
0.09	26.2		{}		-129
0.10	24.6		{}		-129
0.15	20.9		{}		-129
0.20	16.8		{}		-129
0.25	13.5		{}		-129
0.30	11.8		{}		-129
0.35	10.1		{}		-129
0.40	9.3		{}		-129
0.45	8.6		{}		-129
0.50	7.9		{}		-129
0.55	7.4		{}		-129
0.60	6.9		{}		-129
0.65	6.6		{}		-129
0.70	6.5		{}		-129
0.75	6.5		{}		-129
0.80	6.5		{}		-129
0.85	6.5		{}		-129
0.90	6.5		{}		-129
0.95	6.5		{}		-129
1.00	6.5		{}		-129
	};
	\addlegendentry{CiteSeer - \ourattack{}}
	
	\addplot[dashed, color=red, mark options={solid, fill=red, draw=black}, line width=0.5pt, mark size=\marksize, mark=triangle*, visualization depends on=\thisrow{alignment} \as \alignment, nodes near coords, point meta=explicit symbolic,
    every node near coord/.style={anchor=\alignment, font=\scriptsize}] 
table [meta index=2]  {
x   y       label   alignment
0.00    68.5		{}		-129
0.01	33.2		{}		-129
0.02	23.6		{}		-129
0.03	17.8		{}		-129
0.04	15.3		{}		-129
0.05	12.9		{}		-129
0.06	10.9		{}		-129
0.07	9.4		{}		-129
0.08	8.4		{}		-129
0.09	7.4		{}		-129
0.10	6.8		{}		-129
0.15	4.5		{}		-129
0.20	2.6		{}		-129
0.25	1.7		{}		-129
0.30	1.1		{}		-129
0.35	1.0		{}		-129
0.40	0.9		{}		-129
0.45	0.8		{}		-129
0.50	0.8		{}		-129
	};
	\addlegendentry{CiteSeer - \ourattack{}+\emph{Topology}}
	    
    \addplot[color=blue, mark options={solid, fill=blue, draw=black}, line width=0.5pt, mark size=\marksize, mark=diamond*, visualization depends on=\thisrow{alignment} \as \alignment, nodes near coords, point meta=explicit symbolic,
    every node near coord/.style={anchor=\alignment, font=\scriptsize}] 
table [meta index=2]  {
x   y       label   alignment
0.00    78.5		{}		-129
0.05	74.3		{}		-129
0.10	71.2		{}		-129
0.15	68.9		{}		-129
0.20	67.4		{}		-129
0.25	66.2		{}		-129
0.30	65.4		{}		-129
0.35	64.4		{}		-129
0.40	63.7		{}		-129
0.45	63.3		{}		-129
0.50	62.9		{}		-129
	};
	\addlegendentry{PubMed - \ourattack{}}
	
	\addplot[dashed, color=blue, mark options={solid, fill=blue, draw=black}, line width=0.5pt, mark size=\marksize, mark=diamond*, visualization depends on=\thisrow{alignment} \as \alignment, nodes near coords, point meta=explicit symbolic,
    every node near coord/.style={anchor=\alignment, font=\scriptsize}] 
table [meta index=2]  {
x   y       label   alignment
0.00    78.5		{}		-129
0.05	66.8		{}		-129
0.10	57.0		{}		-129
0.15	44.5		{}		-129
0.20	37.6		{}		-129
0.25	33.4		{}		-129
0.30	30.7		{}		-129
0.35	28.1		{}		-129
0.40	26.5		{}		-129
0.45	25.1		{}		-129
0.50	24.1		{}		-129
	};
	\addlegendentry{PubMed - \ourattack{}+\emph{Topology}}
    
	\end{axis}
\end{tikzpicture}
\caption{Test accuracy under our \ourattack{} attack, compared to $\epsilon_0$, the maximal ratio of perturbed features.}
\label{fig:limit-l0}
\end{figure}

\subsection{Sensitivity to $\eps_0$}
\label{subsec:limit-l0}

In \cref{subsec:sensitivity}, we analyzed the effect of $\epsilon_{\infty}$, the maximal allowed perturbation in each vector attribute, on the performance of the attack.
In Cora and CiteSeer, the input features are discrete (i.e., the given input node vector is many-hot).
In such datasets, the interesting analysis focuses on the value of $\epsilon_{0}$, the maximal \emph{fraction} of allowed  perturbed vector elements, on the performance of the attack. Here, we vary the value of $\epsilon_{0}$ across different datasets.
We also included an analysis of $\epsilon_{0}$ for the PubMed dataset, with a constant $\epsilon_\infty=0.04$.
We experimented with limiting $\epsilon_{0}$ and measuring the resulting test accuracy for both \ourattack{} and \ourattack{}+\emph{Topology}. The results appear in \cref{fig:limit-l0}.

As we increase $\epsilon_0$, the test accuracy naturally decreases for all datasets, whether they are discrete or continuous and for both \ourattack{} and \ourattack{}+\emph{Topology}.
\section{Conclusion}
\label{sec:conclusion}
In this paper, we show that GNNs are susceptible to the extremely limited scenario of a \emph{Single-node INdirect Gradient adversariaL Evasion}  (\ourattack{}) attack. \rev{Furthermore, we show that even robustly optimized GNNs and adversarial training fail in defending against a limited single-node attack. We also present a new \emph{Single-Edge gradient adversariaL evasion}  (\ouredgeattack{}) attack that is stronger than its predecessors}.%

\rev{We perform a thorough experimental evaluation across multiple variations of the \ourattack{} attack, datasets, GNN types and also an extensive ablation study.} The practical consequences of these findings are that a single attacker can force a GNN to classify any other node as the attacker's chosen label by slightly perturbing some of the attacker's features.
Furthermore, if the attacker can choose its attacker node, the effectiveness of the attack significantly increases.

We believe that this work will drive research toward exploring novel defense approaches for GNNs. Such defenses can be crucial for real-world systems that are modeled using GNNs.  
We also believe that this work's surprising results motivate a search for a better theoretical understanding of the expressiveness and generalization of GNNs.

\bibliography{single_aaai}
\clearpage
\appendix

\setcounter{figure}{0}  
\setcounter{table}{0}
\setcounter{equation}{0}

\section{Dataset Statistics}
\label{subsec:stats}
Statistics of the datasets are shown in \cref{tab:datasets}.
\begin{table*}
    \centering
      \caption{Dataset statistics.}
    \resizebox{\linewidth}{!}{
    \begin{tabular}{lrrrrrr}
    
        \toprule
         &  \textbf{\#Training} &  \textbf{\#Val}  &  \textbf{\#Test} & \textbf{\#Unlabeled Nodes} & \textbf{\#Classes} & \textbf{Avg.\ Node Degree} \\
          \midrule
         Cora & 140 & 500 & 1000 & 2708 & 7 & 3.9\\
         CiteSeer & 120 & 500 & 1000 & 3327 & 6 & 2.7\\
         PubMed & 60 & 500 &1000 & 19717 & 3 & 4.5\\
         Twitter & 4474 & 248 & 249 & 95415 & 2 & 45.6\\
	    \bottomrule
    \end{tabular}
    }
    \label{tab:datasets}
\end{table*} 

\rev{\cref{tab:L_0andL_inf} displays the default settings for the $\ell_0$ norm and $\ell_\infty$ norm of each dataset, where the $\ell_0$ and $\ell_\infty$ norms are influenced by the average ratio of non-zero attributes and the average amplitude of the non-zero attributes, respectively.}

\begin{table*}
    \centering
      \caption{Default $\ell_0$ and $\ell_\infty$.}
      \resizebox{\linewidth}{!}{
    \begin{tabular}{lrrrrrr}
        \toprule
         &  \textbf{\#Features} &  \textbf{Avg.\ ratio of non-zero features}  & \textbf{$\ell_0$} & \textbf{Avg.\ Amplitude of non-zero feature} & \textbf{$\ell_\infty$}\\
          \midrule
         Cora & 1433 & 0.013 & 0.01 & - & -\\
         CiteSeer & 3703 & 0.009 & 0.01 & - & -\\
         PubMed & 500 & 0.1 & 0.05 & 0.04 & 0.04\\
         Twitter & 10000 & 0.052 & 0.1 & 0.0009 & 0.01\\
	    \bottomrule
    \end{tabular}
    }
    \label{tab:L_0andL_inf}
\end{table*}

\subsection{Additional GNN Types}
\label{subsec:gnntypes}
\cref{tab:gat-untargeted,tab:gin-untargeted,tab:sage-untargeted} present the test accuracy of different attacks applied on GAT \cite{velic2018graph}, GIN \cite{xu2018powerful}, GraphSAGE \cite{hamilton2017inductive}, and SGC \cite{wu2019simplifying}, showing the effectiveness of \ourattack{} across different GNN types.
\begin{table}[h!]
    \centering
    \begin{tabular}{lrrr}
        \toprule
        & \multicolumn{1}{c}{\textbf{Cora}} 	& \multicolumn{1}{c}{\textbf{CiteSeer}} & \multicolumn{1}{c}{\textbf{PubMed}} \\% & \multicolumn{1}{c}{\textbf{Twitter}}
        \midrule
        Clean & 78.2\eb{1.4}  & 65.6\eb{1.4} & 75.0\eb{0.3}\\
        \ourattack{}  & 67.9\eb{1.1} & 45.4\eb{3.4} & 68.9\eb{3.7} \\
        \midrule
        \ourattack{}+\emph{GradChoice} & 58.2\eb{1.6} & 37.0\eb{4.7} & 52.0\eb{2.1}\\
        \ourattack{}+\emph{Topology} & 59.4\eb{2.2} & 36.6\eb{4.3} & 51.5\eb{5.5}\\
        \midrule
        \ourattack{}-\emph{hops} & 72.4\eb{0.9} & 54.7\eb{3.0} & 70.5\eb{3.5}\\
        \ourattack{}-\emph{two attackers} & 64.4\eb{1.0} & 41.2\eb{3.9} & 62.6\eb{4.9}\\
        \ourattack{}-\emph{direct} & 55.4\eb{3.7} & 31.1\eb{3.9} & 47.7\eb{6.9}\\
        \midrule
        \ouredgeattack{}  & 70.5\eb{0.6}  & 49.4\eb{1.4}  & 64.9\eb{1.0}\\
        \ouredgeattack{}+\emph{GradChoice} & 67.8\eb{4.9} & 48.3\eb{5.1} & 63.5\eb{4.6}\\% & 82.7\eb{0.0}\\
        \bottomrule
    \end{tabular}
    \caption{
    Test accuracy of GAT under different non-targeted attacks}
    \label{tab:gat-untargeted}
\end{table} 

\begin{table}[h!]
    \centering
    \begin{tabular}{lrrrl}
        \toprule
        & \multicolumn{1}{c}{\textbf{Cora}} 	& \multicolumn{1}{c}{\textbf{CiteSeer}} & \multicolumn{1}{c}{\textbf{PubMed}} \\
		\midrule
		Clean & 57.7\eb{1.4} & 39.5\eb{1.9} & 55.0\eb{4.4}\\
        \ourattack{}  & 14.9\eb{1.8} & 14.9\eb{1.0} & 36.8\eb{8.1}\\
        \midrule
        \ourattack{}+\emph{GradChoice} & 26.8\eb{1.7} & 8.8\eb{1.1} & 24.9\eb{7.3}\\
        \ourattack{}+\emph{Topology} & 25.9\eb{1.7} & 8.1\eb{1.5} & 21.2\eb{7.2}\\
        \midrule
        \ourattack{}-\emph{hops} & 42.1\eb{1.9} & 18.9\eb{0.7} & 37.7\eb{8.2}\\
        \ourattack{}-\emph{two attackers} & 32.5\eb{2.0} & 11.9\eb{1.4} & 31.1\eb{8.3}\\
        \ourattack{}-\emph{direct} & 20.4\eb{1.9} & 5.2\eb{2.0} & 41.1\eb{0.7}\\
        \midrule
        \ouredgeattack{}  & 32.9\eb{3.1} & 18.5\eb{3.0} & 33.3\eb{1.7}\\
        \ouredgeattack{}+\emph{GradChoice} & 10.7\eb{2.8} & 4.8\eb{2.1} & 10.3\eb{1.0}\\
        \bottomrule
    \end{tabular}
    \caption{
    Test accuracy of GIN under different non-targeted attacks}
    \label{tab:gin-untargeted}
\end{table} 

\begin{table}[h!]
    \centering
    \begin{tabular}{lrrrl}
        \toprule
        & \multicolumn{1}{c}{\textbf{Cora}} 	& \multicolumn{1}{c}{\textbf{CiteSeer}} & \multicolumn{1}{c}{\textbf{PubMed}} \\
		\midrule
		Clean & 78.7\eb{0.3} & 66.0\eb{0.5} & 75.9\eb{0.5} \\
        \ourattack{}  & 70.7\eb{2.0} & 44.7\eb{2.3} & 72.2\eb{0.5}\\
        \midrule
        \ourattack{}+\emph{GradChoice} & 56.2\eb{2.9} & 28.1\eb{1.7} & 46.5\eb{0.4}\\
        \ourattack{}+\emph{Topology} & 57.5\eb{2.9} & 29.8\eb{1.9} & 43.6\eb{0.5}\\
        \midrule
        \ourattack{}-\emph{hops} & 76.3\eb{1.9} & 60.2\eb{2.6} & 72.2\eb{0.5}\\
        \ourattack{}-\emph{two attackers} & 65.8\eb{1.7} & 40.8\eb{1.5} & 67.6\eb{0.7}\\
        \ourattack{}-\emph{direct} & 47.7\eb{1.4} & 21.9\eb{2.0} & 41.1\eb{0.7}\\
        \midrule
        \ouredgeattack{}  & 62.9\eb{1.9} & 45.9\eb{3.4} & 64.2\eb{1.6} \\
        \ouredgeattack{}+\emph{GradChoice} & 48.9\eb{2.7} & 40.4\eb{3.3} & 64.7\eb{1.1}  \\
        \bottomrule
    \end{tabular}
    \caption{
     Test accuracy of GraphSAGE under different non-targeted attacks}
    \label{tab:sage-untargeted}
\end{table} 

\begin{table}[h!]
    \centering
    \begin{tabular}{lrrr}
        \toprule & \multicolumn{1}{c}{\textbf{Cora}} & \multicolumn{1}{c}{\textbf{CiteSeer}}\\
		\midrule
		Clean & 79.9\eb{0.6} & 67.9\eb{0.2} \\
        \ourattack{}  & 70.4\eb{0.3} & 46.6\eb{0.5} \\
        \midrule
        \ourattack{}+\emph{GradChoice} & 60.4\eb{0.5} & 35.8\eb{0.4}\\
        \ourattack{}+\emph{Topology} & 55.9\eb{0.5} & 33.4\eb{0.4}\\
        \midrule
        \ourattack{}-\emph{hops} & 75.9\eb{0.3} & 59.5\eb{0.4} \\
        \ourattack{}-\emph{two attackers} & 64.5\eb{0.5} & 40.8\eb{0.5}\\
        \ourattack{}-\emph{direct} & 45.1\eb{0.4} & 22.6\eb{0.5}\\
        \midrule
		\ouredgeattack{}  & 71.3\eb{1.0} & 55.0\eb{1.6} \\
        \ouredgeattack{}+\emph{GradChoice} & 29.7\eb{1.7} & 13.5\eb{2.0}\\
        \bottomrule
    \end{tabular}
    \caption{
     Test accuracy of SGC \cite{wu2019simplifying} under different non-targeted attacks.}
    \label{tab:SGC-untargeted_Cora}
\end{table} 

\section{Targeted Attacks}
\label{subsec:targeted}
\begin{table}[h!]
    \centering
    \begin{tabular}{lrrrl}
        \toprule
        & \multicolumn{1}{c}{\textbf{Cora}} 	& \multicolumn{1}{c}{\textbf{CiteSeer}} & \multicolumn{1}{c}{\textbf{PubMed}}\\
		\midrule
        \ourattack{}  & 9.1\eb{0.4} & 21.6\eb{0.8} & 12.7\eb{0.8}\\
        \ourattack{}+\emph{GradChoice} & 14.5\eb{1.2} & 27.9\eb{1.1} & 15.9\eb{0.8}\\
        \ourattack{}+\emph{Topology} & 17.0\eb{1.5} & 30.9\eb{1.0} & 17.2\eb{0.7}\\
        \ourattack{}-\emph{hops} & 4.2\eb{0.3} & 8.8\eb{1.0} & 12.4\eb{0.9}\\
        \midrule
        \ouredgeattack{}  & 8.0\eb{0.7}  & 14.8\eb{0.5}  & 20.1\eb{0.6}\\
        \ouredgeattack{}+\emph{GradChoice} & 59.4\eb{0.9} & 78.7\eb{0.9} & 80.1\eb{0.6}\\
        \bottomrule
    \end{tabular}
    \caption{
    Success rate (higher is better) of different \emph{targeted} attacks on a GCN network.}
    \label{tab:targeted-gcn}
\end{table} 

\cref{tab:targeted-gcn,tab:targeted-gin,tab:targeted-gat,tab:targeted-sage} show the results of \emph{targeted} attacks across datasets and approaches. Differently from other tables that show test accuracy, \cref{tab:targeted-gcn,tab:targeted-gin,tab:targeted-gat,tab:targeted-sage} present the targeted attack's \emph{success rate}, which is the fraction of test examples that the attack managed to force to make \emph{a specific label prediction} (in these results, higher is better). %

\begin{table}[h!]
    \centering
    \begin{tabular}{lrrrl}
        \toprule
        & \multicolumn{1}{c}{\textbf{Cora}} 	& \multicolumn{1}{c}{\textbf{CiteSeer}} & \multicolumn{1}{c}{\textbf{PubMed}}\\
		\midrule
        \ourattack{}  & 13.7\eb{1.2} & 28.7\eb{3.6} & 15.5\eb{1.7}\\
        \ourattack{}+\emph{GradChoice} & 25.8\eb{2.2} & 38.6\eb{4.7} & 20.6\eb{2.3}\\
        \ourattack{}+\emph{Topology} & 24.9\eb{2.6} & 18.4\eb{5.3} & 22.1\eb{1.6}\\
        \ourattack{}-\emph{hops} & 7.2\eb{0.9} & 14.4\eb{2.0} & 14.7\eb{1.8}\\
        \midrule
        \ouredgeattack{}  & 6.1\eb{0.4}  & 12.5\eb{1.2}  & 17.9\eb{1.5}\\
        \ouredgeattack{}+\emph{GradChoice} & 6.0\eb{1.4} & 14.6\eb{2.8} & 22.3\eb{3.6}\\
        \bottomrule
    \end{tabular}
    \caption{
    Success rate (higher is better)  of different \emph{targeted} attacks on a GAT network.}
    \label{tab:targeted-gat}
\end{table} 

\begin{table}[h!]
    \centering
    \begin{tabular}{lrrrl}
        \toprule
        & \multicolumn{1}{c}{\textbf{Cora}} 	& \multicolumn{1}{c}{\textbf{CiteSeer}} & \multicolumn{1}{c}{\textbf{PubMed}}\\
		\midrule
        \ourattack{}  & 23.6\eb{0.4} & 50.9\eb{4.5} & 39.4\eb{9.2}\\
        \ourattack{}+\emph{GradChoice} & 36.0\eb{0.7} & 64.5\eb{5.0} & 49.2\eb{7.2}\\
        \ourattack{}+\emph{Topology} & 36.9\eb{1.3} & 64.8\eb{5.5} & 52.2\eb{7.1}\\
        \ourattack{}-\emph{hops} & 17.1\eb{1.0} & 34.9\eb{3.3} & 38.5\eb{9.3}\\
        \midrule
        \ouredgeattack{}  & 16.8\eb{1.2} & 25.6\eb{1.0} & 37.9\eb{2.6} \\
        \ouredgeattack{}+\emph{GradChoice} & 44.7\eb{4.7} & 55.0\eb{7.0} & 64.9\eb{11.8} \\
        \bottomrule
    \end{tabular}
    \caption{
    Success rate (higher is better)  of different \emph{targeted} attacks on a GIN network.}
    \label{tab:targeted-gin}
\end{table} 

\begin{table}[h!]
    \centering
    \begin{tabular}{lrrrl}
        \toprule
        & \multicolumn{1}{c}{\textbf{Cora}} 	& \multicolumn{1}{c}{\textbf{CiteSeer}} & \multicolumn{1}{c}{\textbf{PubMed}}\\
		\midrule
        \ourattack{}  & 9.8\eb{0.7} & 25.2\eb{1.1} & 14.2\eb{0.9}\\
        \ourattack{}+\emph{GradChoice} & 20.5\eb{2.1} & 40.5\eb{1.5} & 23.3\eb{1.4}\\
        \ourattack{}+\emph{Topology} & 20.1\eb{1.4} & 39.1\eb{1.6} & 26.8\eb{1.1}\\
        \ourattack{}-\emph{hops} & 4.6\eb{0.6} & 8.6\eb{0.7} & 12.9\eb{0.9}\\
        \midrule
        \ouredgeattack{}  & 7.6\eb{0.3} & 16.3\eb{1.7} & 19.1\eb{1.4} \\
        \ouredgeattack{}+\emph{GradChoice} & 9.3\eb{0.9} & 14.7\eb{1.1} & 19.6\eb{0.8} \\
        \bottomrule
    \end{tabular}
    \caption{
    Success rate (higher is better)  of different \emph{targeted} attacks on a GraphSAGE network.}
    \label{tab:targeted-sage}
\end{table} 

\subsection{Adversarial Training}
\label{subsec:appenidx-adversarial}
\begin{table}
    \centering
    \begin{tabular}{lrrrr}
        \toprule
        & \textbf{Std.} & 
        \textbf{Adv.} & 
        \textbf{Adv.+Top.}\\
        \midrule
        Clean (no attack) & 78.5\eb{0.6} & 76.0\eb{1.8} & 76.0\eb{1.9}\\
        \ourattack{}  & 74.3\eb{0.3} & 73.5\eb{1.9} & 72.9\eb{1.6}\\
        \ourattack{}-\emph{hops} & 75.5\eb{0.2} & 73.5\eb{1.8} & 73.7\eb{1.9}\\
        \midrule
        \ourattack{}+\emph{GradChoice} & 68.5\eb{0.1} & 66.5\eb{2.6} & 66.5\eb{2.7}\\
        \ourattack{}+\emph{Topology} & 66.0\eb{0.5} & 64.8\eb{1.9} & 64.7\eb{1.8}\\
        \midrule
        \ourattack{}-\emph{two attackers} & 69.9\eb{0.3} & 69.1\eb{1.6} & 69.0\eb{1.9}\\
        \ourattack{}-\emph{direct} & 46.3\eb{1.1} & 47.1\eb{0.7} & 46.8\eb{0.6}\\
        \bottomrule
    \end{tabular}
    \caption{Test accuracy on PubMed of a model that was trained adversarially on the \ourattack{} attack or the \ourattack{}+\emph{Topology} attack. }
    \label{tab:adversarial_model}
\end{table} 
\rev{\cref{tab:adversarial_model} presents the test accuracy on PubMed of a model that was trained adversarially on the \ourattack{} attack or the \ourattack{}+\emph{Topology} attack. It shows that adversarial training does not improve the model robustness against the different \ourattack{} attacks.}

\section{Additional Robust GNN Types}
\label{subsec:additional-robust}
\begin{table*}
    \centering
\resizebox{\linewidth}{!}{    
    \begin{tabular}{lrrrrr}
        \toprule 
        & \multicolumn{2}{c}{GAL} (unique train/val/test split) & \multicolumn{3}{c}{LAT-GCN}\\
        & \textbf{CiteSeer} & \textbf{PubMed} & \textbf{Cora} & \textbf{CiteSeer} & \textbf{PubMed}\\
		\midrule
		Clean & 78.4\eb{1.3} & 74.9\eb{3.5} & 80.1\eb{0.3} & 69.4\eb{0.5} & 73.7\eb{1.8}\\
        \ourattack{}  & 25.1\eb{2.9} & 71.0\eb{2.4} & 62.2\eb{0.8} & 35.2\eb{0.7} & 43.8\eb{5.4}\\
        \midrule
        \ourattack{}+\emph{GradChoice} & 10.1\eb{2.8} & 63.7\eb{2.7} & 55.3\eb{0.8} & 27.5\eb{1.1} & 40.0\eb{6.9}\\
        \ourattack{}+\emph{Topology} &  10.5\eb{2.0} & 60.7\eb{1.8} & 52.9\eb{0.7} & 27.0\eb{0.5} & 40.1\eb{6.9}\\
        \midrule
        \ourattack{}-\emph{hops} & 42.3\eb{4.0} & 72.5\eb{2.4} & 67.0\eb{0.8} & 46.5\eb{0.6} & 44.9\eb{5.2}\\
        \ourattack{}-\emph{two attackers} & 26.1\eb{2.9} & 67.2\eb{2.4} & 57.1\eb{0.5} & 34.0\eb{0.9} & 40.5\eb{5.7}\\
        \ourattack{}-\emph{direct} & 7.9\eb{2.3} & 54.5\eb{1.7} & 44.8\eb{0.8} & 17.9\eb{1.3} & 31.0\eb{5.0}\\
        \bottomrule
    \end{tabular}
    }
    \caption{
     Test accuracy of Graph Adversarial Networks (GAL) \cite{liao2021information} with a unique unique train/val/test split and LAT-GCN \cite{jin2020robust} under different non-targeted attacks}
    \label{tab:robust-untargeted-2}
\end{table*} 
\rev{\cref{tab:robust-untargeted-2} presents the test accuracy of different attacks applied on 
\emph{GAL} \cite{liao2021information} and \emph{LAT-GCN} \cite{jin2020robust}. It shows the effectiveness of \ourattack{} across different robust GNN types.}

\subsection{The Trade-Off Between $\eps_0$ and $\eps_\infty$ }
\label{subsec:limit-l0-sensitivity}

\begin{figure}
\centering
        \includegraphics[width=\linewidth]{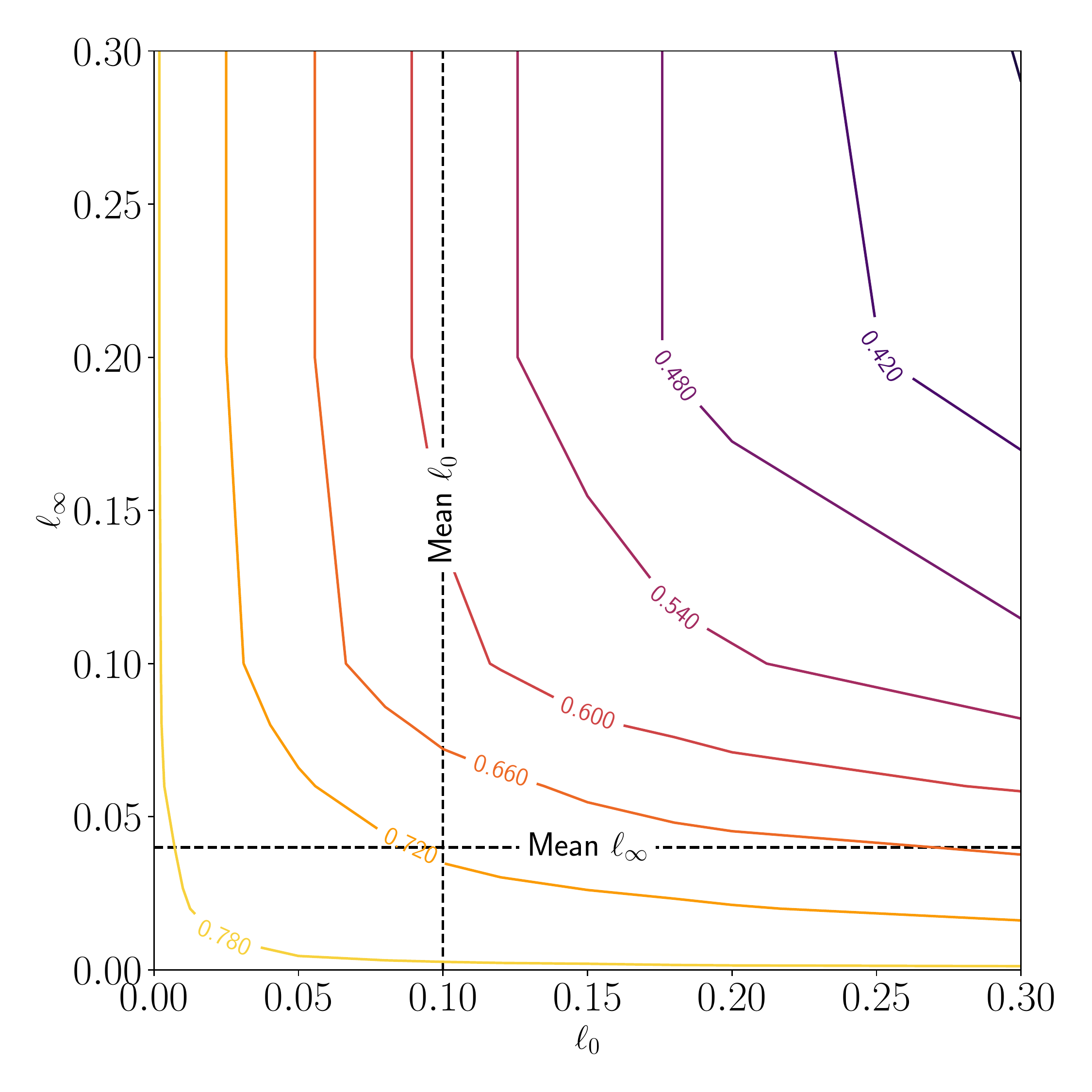}
 \caption{Constant accuracy lines for $\epsilon_{0}$ and $\epsilon_\infty$, interpolated based on grid with $\epsilon_0 \in \qty{0, 0.01, 0.02, 0.05, 0.08, 0.1, 0.12, 0.15, 0.18, 0.2, 0.3}$ and $\epsilon_\infty \in \qty{0,0.01, 0.02, 0.04, 0.06, 0.08, 0.1, 0.15, 0.2, 0.3}$.}
\label{fig:contour}
\end{figure}

\cref{fig:contour} shows a contour plot of accuracy as a function of $\epsilon_{\infty}$ and $\epsilon_{0}$. We attacked the PubMed dataset with out \ourattack{} with $\epsilon_0 \in \qty{0, 0.01, 0.02, 0.05, 0.08, 0.1, 0.12, 0.15, 0.18, 0.2, 0.3}$ and $\epsilon_\infty \in \qty{0,0.01, 0.02, 0.04, 0.06, 0.08, 0.1, 0.15, 0.2, 0.3}$.
These contours detail the trade-off between $\epsilon_{\infty}$ and $\epsilon_{0}$ values for a constant desired accuracy. %

\section{Additional Baselines}
\subsection{Zero-Features Approach}
We experimented with an additional baseline where we set $\eta=-x_a$ as the feature perturbation. The objective of experimenting with such an attack is to illustrate that \ourattack{} can find better perturbations than simply canceling the node feature vector, making the new vector a vector of zeros. %

\begin{table}[h!]
    \centering
    
    \begin{tabular}{lr}
        \toprule
        & \multicolumn{1}{c}{\textbf{PubMed}}\\
        \midrule
        Clean & 78.5\eb{0.6} \\
        
        \ourattack{} & 74.3\eb{0.3} \\
        Zero features & 77.8\eb{0.1} \\
        \bottomrule
    \end{tabular}
    \caption{
    Test accuracy of our zero features attack on a GCN network.}
    \label{tab:zero-features}
\end{table} 

As shown, \emph{zero features} is barely effective (compared to ``Clean''), and  \ourattack{} can find much better perturbations.

\subsection{Injection attacks}
We also studied an additional type of a realistic attack that is based on node injection.
In this approach, we inserted a new node into the graph with a single edge attached to our victim node. The attack was performed by perturbing the injected node's attributes. 
This attack is very powerful, reducing the test accuracy to 10.1\eb{0.9}\% on PubMed.

\section{Distance Between Attacker and Victim}
\label{subsec:distance}
In \cref{subsec:results}, we found that \ourattack{} performs similarly to \ourattack{}-\emph{hops}, although \ourattack{}-\emph{hops} samples  attacker node $a$ whose distance from victim node $v$ is at least 2.
We further question whether the effectiveness of the attack depends on the distance between the attacker and the victim. We trained a new model for each dataset using $L=8$ layers.
Then, for each test victim node, we sampled attackers according to their distance to the test node.

As shown in \cref{fig:distance}, the effectiveness of the attack \emph{increases} as the distance between the attacker and the victim \emph{decreases}. 
At distance of 4, the test accuracy saturates. %
A possible explanation is that apparently more than a few layers (e.g., $L=2$ in \citet{kipf2016semi}) are not needed in most datasets. Thus, the rest of the layers can theoretically learn \emph{not} to pass much of their input, starting from the redundant layers, excluding adversarial signals as well.

\renewcommand{\marksize}{2pt}

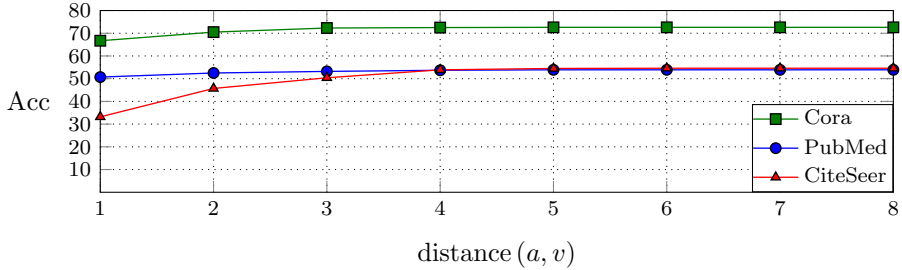
\begin{figure}
\centering
	\begin{tikzpicture}[scale=1]
	\begin{axis}[
		xlabel={$\text{distance}\left(a,v\right)$},
		ylabel={Acc},
		y label style={at={(axis description cs:0.25,0.5)},anchor=east},
		ylabel near ticks,
        legend style={at={(1,0)},anchor=south east,
        font=\footnotesize,inner xsep=0pt, inner ysep=0pt},
        legend cell align={left},
        xmin=1, xmax=8,
        ymin=0.0, ymax=80.0,
        xtick={1,2,3,4,5,6,7,8},
        ytick={10,20,...,100},
        ylabel style={rotate=-90}, %
        tick label style={font=\footnotesize} ,
        grid = major,
        major grid style={dotted,black},
        height = 4cm,
        width = 1\linewidth,
    ]

\addplot[color=ao, mark options={solid, fill=ao, draw=black}, mark=square*, line width=0.5pt, mark size=\marksize, visualization depends on=\thisrow{alignment} \as \alignment, nodes near coords, point meta=explicit symbolic,
    every node near coord/.style={anchor=\alignment, font=\scriptsize}] 
table [meta index=2]  {
x   y       label   alignment
1	66.7		{}		-129
2	70.5		{}		-129
3	72.3		{}		-129
4	72.5		{}		-129
5	72.6		{}		-129
6	72.6		{}		-129
7	72.6		{}		-129
8   72.6		{}		-129
	};
    \addlegendentry{Cora}

\addplot[color=blue, mark options={solid, fill=blue, draw=black}, mark=*, line width=0.5pt, mark size=\marksize, visualization depends on=\thisrow{alignment} \as \alignment, nodes near coords, point meta=explicit symbolic,
    every node near coord/.style={anchor=\alignment, font=\scriptsize}] 
table [meta index=2]  {
x   y       label   alignment
1	50.7		{}		-129
2	52.5		{}		-129
3	53.2		{}		-129
4	53.7		{}		-129
5	53.9		{}		-129
6	53.9		{}		-129
7	53.9		{}		-129
8	53.9		{}		-129
	};
    \addlegendentry{PubMed}

    \addplot[color=red, mark options={solid, fill=red, draw=black}, mark=triangle*, line width=0.5pt, mark size=\marksize, visualization depends on=\thisrow{alignment} \as \alignment, nodes near coords, point meta=explicit symbolic,
    every node near coord/.style={anchor=\alignment, font=\scriptsize}] 
table [meta index=2]  {
x   y       label   alignment
1	33.2		{}		-129
2	45.7		{}		-129
3	50.4		{}		-129
4	53.9		{}		-129
5	54.5		{}		-129
6	54.6		{}		-129
7	54.6		{}		-129
8	54.6		{}		-129
	};
	\addlegendentry{CiteSeer}
    
	\end{axis}
\end{tikzpicture}
\caption{Test accuracy compared to the distance between the attacker and the victim, when trained with $L=8$ layers.}
\label{fig:distance}
\end{figure}

\section{Larger number of attackers}
\label{subsubsec:num-attackers}

\begin{table}
    \centering
    
    \begin{tabular}{cr}
        \toprule
        \textbf{Number of} & \multirow{2}{*}{\textbf{PubMed}}\\
        \textbf{attackers} &\\
        \midrule
        1 & 74.3\eb{0.3} \\
        2 & 69.9\eb{0.3} \\
        3 & 66.9\eb{0.4} \\
        4 & 63.2\eb{0.9} \\
        5 & 59.7\eb{0.9} \\
        \bottomrule
    \end{tabular}
    \caption{
    Test accuracy for \ourattack{} with a different number of attackers on PubMed.}
    \label{tab:attacker-num}
\end{table} 

We performed additional experiments with up to five randomly sampled attacker nodes simultaneously and perturb their features using the same approach as \ourattack{} (\cref{tab:attacker-num}). As expected, allowing a larger number of attackers reduces the test accuracy. The main observation in this paper, however, is that even a single attacker node is surprisingly effective.

\section{\emph{Multi-Edge} attacks}
\label{subsec:multiedge}
We strengthened our \ouredgeattack{} attack by allowing it to add and remove multiple edges that are connected to the attacker node, thereby creating \emph{Multi-Edge}. Accordingly,  \emph{Multi-Edge+GradChoice} adds and removes multiple edges from the entire graph. 

\begin{table}[h!]
    \centering
    
    \begin{tabular}{lrrrl}
        \toprule & \multicolumn{1}{c}{\textbf{PubMed}}\\
		\midrule
		Clean & 78.5\eb{0.6}\\
		\midrule
		\ouredgeattack{}  & 65.1\eb{1.3}\\
		\emph{Multi-Edge} & 64.5\eb{0.2}\\
        \midrule
        \ouredgeattack{}+\emph{GradChoice} & 15.3\eb{0.4}\\
        \emph{Multi-Edge+GradChoice} & 15.3\eb{0.5}\\
        \bottomrule
    \end{tabular}
    \caption{
    Test accuracy of GCN using Multi-Edge attacks}
    \label{tab:MultiEdge}
\end{table} 

As shown in \cref{tab:MultiEdge}, allowing the attacker node to add and remove multiple edges (\emph{Multi-Edge} and \emph{Multi-Edge+GradChoice}) results in a very minor improvement compared to \ouredgeattack{} and \ouredgeattack{}+\emph{GradChoice}.%

\end{document}